# An Analysis of Reduced Error Pruning

**Tapio Elomaa** ELOMAA@CS.HELSINKI.FI
**Matti Kääriäinen** MATTI.KAARIAINEN@CS.HELSINKI.FI
*Department of Computer Science*
*P. O. Box 26 (Teollisuuskatu 23)*
*FIN-00014 University of Helsinki, Finland*

## Abstract

Top-down induction of decision trees has been observed to suffer from the inadequate functioning of the pruning phase. In particular, it is known that the size of the resulting tree grows linearly with the sample size, even though the accuracy of the tree does not improve. Reduced Error Pruning is an algorithm that has been used as a representative technique in attempts to explain the problems of decision tree learning.

In this paper we present analyses of Reduced Error Pruning in three different settings. First we study the basic algorithmic properties of the method, properties that hold independent of the input decision tree and pruning examples. Then we examine a situation that intuitively should lead to the subtree under consideration to be replaced by a leaf node, one in which the class label and attribute values of the pruning examples are independent of each other. This analysis is conducted under two different assumptions. The general analysis shows that the pruning probability of a node fitting pure noise is bounded by a function that decreases exponentially as the size of the tree grows. In a specific analysis we assume that the examples are distributed uniformly to the tree. This assumption lets us approximate the number of subtrees that are pruned because they do not receive any pruning examples.

This paper clarifies the different variants of the Reduced Error Pruning algorithm, brings new insight to its algorithmic properties, analyses the algorithm with less imposed assumptions than before, and includes the previously overlooked empty subtrees to the analysis.

## 1. Introduction

Decision tree learning is usually a two-phase process (Breiman, Friedman, Olshen, & Stone, 1984; Quinlan, 1993). First a tree reflecting the given sample as faithfully as possible is constructed. If no noise prevails, the accuracy of the tree is perfect on the *training examples* that were used to build the tree. In practice, however, the data tends to be noisy, which may introduce contradicting examples to the training set. Hence, 100% accuracy cannot necessarily be obtained even on the training set. In any case, the resulting decision tree is *overfitted* to the sample; in addition to the general trends of the data, it encodes the peculiarities and particularities of the training data, which makes it a poor predictor of the class label of future instances. In the second phase of induction, the decision tree is *pruned* in order to reduce its dependency on the training data. Pruning aims at removing from the tree those parts that are likely to only be due to the chance properties of the training set.

The problems of the two-phased top-down induction of decision trees are well-known and have been extensively reported (Catlett, 1991; Oates & Jensen, 1997, 1998). The size of the tree grows linearly with the size of the training set, even though after a while no accuracy is gained through the increased tree complexity. Obviously, pruning is intended to fight this effect. Another defect is observed when the data contains no relevant attributes; i.e., when the class labels of the examples are independent of their attribute values. Clearly, a single-node tree predicting the majority label of the examples should result in this case, since no help can be obtained by querying the attribute values. In practice, though, often large decision trees are built from such data.

Many alternative pruning schemes exist (Mingers, 1989a; Esposito, Malerba, & Semeraro, 1997; Frank, 2000). They differ, e.g., on whether a single pruned tree or a series of pruned trees is produced, whether a separate set of *pruning examples* is used, which aspects (classification error and tree complexity) are taken into account in pruning decisions, how these aspects are determined, and whether a single scan through the tree suffices or whether iterative processing is required. The basic pruning operation that is applied to the tree is the replacement of an internal node together with the subtree rooted at it with a leaf. Also more elaborated tree restructuring operations are used by some pruning techniques (Quinlan, 1987, 1993). In this paper, the only pruning operation that is considered is the replacement of a subtree by the *majority leaf*, i.e., a leaf labeled by the majority class of the examples reaching it. Hence, a *pruning of a tree* is a subtree of the original tree with just zero, one, or more internal nodes changed into leaves.

*Reduced Error Pruning* (subsequently REP for short) was introduced by Quinlan (1987) in the context of decision tree learning. It has subsequently been adapted to rule set learning as well (Pagallo & Haussler, 1990; Cohen, 1993). REP is one of the simplest pruning strategies. In practical decision tree pruning REP is seldom used, because it has the disadvantage of requiring a separate set of examples for pruning. Moreover, it is considered too aggressive a pruning strategy that *overprunes* the decision tree, deleting relevant parts from it (Quinlan, 1987; Esposito et al., 1997). The need for a pruning set is often considered harmful because of the scarceness of the data. However, in the data mining context the examples are often abundant and setting a part of them aside for pruning purposes presents no problem.

Despite its shortcomings REP is a baseline method to which the performance of other pruning algorithms is compared (Mingers, 1989a; Esposito, Malerba, & Semeraro, 1993; Esposito et al., 1997). It presents a good starting point for understanding the strengths and weaknesses of the two-phased decision tree learning and offers insight to decision tree pruning. REP has the advantage of producing the smallest pruning among those that are the most accurate with respect to the pruning set. Recently, Oates and Jensen (1999) analyzed REP in an attempt to explain why and when decision tree pruning fails to control the growth of the tree, even though the data do not warrant the increased size. We approach the same subject, but try to avoid restricting the analysis with unnecessary assumptions. We also consider an explanation for the unwarranted growth of the size of the decision tree.

In this paper we analyze REP in three different settings. First, we explore the basic algorithmic properties of REP, which apply regardless of the distribution of examples presented to the learning algorithm. Second, we study, in a probabilistic setting, the situation in which the attribute values are independent of the classification of an example. Even though this pure noise fitting situation is not expected to arise when the whole pruning set is considered, it is encountered at lower levels of the tree, when all relevant attributes have already been exhausted. We further assume that all subtrees receive at least one pruning example, so that none of them can be directly pruned due to not receiving any examples. The class value is also assigned at random to the pruning examples. In our third analysis it is assumed that each pruning example has an equal chance to end up in any one of the subtrees of the tree being pruned. This rather theoretical setting lets us take into account those subtrees that do not receive any examples. They have been left without attention in earlier analyses.

The rest of this paper is organized as follows. The next section discusses the different versions of the REP algorithm and fixes the one that is analyzed subsequently. In Section 3 we review earlier analyses of REP. Basic algorithmic properties of REP are examined in Section 4. Then, in Section 5, we carry out a probabilistic analysis of REP, without making any assumptions about the distribution of examples. We derive a bound for the pruning probability of a tree which depends exponentially on the relation of the number of pruning examples and the size of the tree. Section 6 presents an analysis, which assumes that the pruning examples distribute uniformly to the subtrees of the tree. This assumption lets us sharpen the preceding analysis on certain aspects. However, the bounds of Section 5 hold with certainty, while those of Section 6 are approximate results. Further related research is briefly reviewed in Section 7 and, finally, in Section 8 we present the concluding remarks of this study.

## 2. Reduced Error Pruning Algorithm

REP was never introduced algorithmically by Quinlan (1987), which is a source of much confusion. Even though REP is considered and appears to be a very simple, almost trivial, algorithm for pruning, there are many different algorithms that go under the same name. No consensus exists whether REP is a bottom-up algorithm or an iterative method. Neither is it obvious whether the training set or pruning set is used to decide the labels of the leaves that result from pruning.

### 2.1 High-Level Control

Quinlan's (1987, p. 225–226) original description of REP does not clearly specify the pruning algorithm and leaves room for interpretation. It includes, e.g., the following characterizations.

> "For every non-leaf subtree $S$ of $T$ we examine the change in misclassifications over the test set that would occur if $S$ were replaced by the best possible leaf. If the new tree would give an equal or fewer number of errors and $S$ contains no subtree with the same property, $S$ is replaced by the leaf. The process continues until any further replacements would increase the number of errors over the test set.
>
> [...] the final tree is the most accurate subtree of the original tree with respect to the test set and the smallest tree with that accuracy."

Quinlan (1987, p. 227) also later continues to give the following description.

> "This method [pessimistic pruning] has two advantages. It is much faster than either of the preceding methods [cost-complexity and reduced error pruning] since each subtree is examined at most once."

On one hand this description requires the nodes to be processed in a bottom-up manner, since subtrees must be checked for "the same property" before pruning a node but, on the other hand, the last quotation would indicate REP to be an iterative method. We take REP to have the following single-scan bottom-up control strategy like in most other studies (Oates & Jensen, 1997, 1998, 1999; Esposito et al., 1993, 1997; Kearns & Mansour, 1998).

> Nodes are pruned in a single bottom-up sweep through the decision tree, pruning each node is considered as it is encountered. The nodes are processed in postorder.

By this order of node processing, any tree that is a candidate for pruning itself cannot contain a subtree that could still be pruned without increasing the tree's error.

Due to the ambiguity of REP's definition, a different version of REP also lives on (Mingers, 1989a; Mitchell, 1997). It is probably due to Mingers' (1989) interpretation of Quinlan's ambiguous definition.

> Nodes are pruned iteratively, always choosing the node whose removal most increases the decision tree accuracy over the pruning set. The process continues until further pruning is harmful.

However, this algorithm appears to be incorrect. Esposito et al. (1993, 1997) have shown that a tree produced by this algorithm does not meet the objective of being the most accurate subtree with respect to the pruning set. Moreover, this algorithm overlooks the explicit requirement of checking whether a subtree would lead to reduction of the classification error.

Other iterative algorithms could be induced from Quinlan's original description. However, if the explicit requirement of checking whether a subtree could be pruned before pruning a supertree is obeyed, then these versions of REP will all reduce to the more efficient bottom-up algorithm.

### 2.2 Leaf Labeling

Another source of confusion in Quinlan's (1987) description of REP is that it is not clearly specified how to choose the labels for the leaves that are introduced to the tree through pruning. Oates and Jensen (1999) interpreted that the intended algorithm would label the new leaves according to the majority class of the *training* examples, but themselves analyzed a version of the algorithm where the new leaves obtain as their labels the majority of the *pruning* examples. Oates and Jensen motivated their choice by the empirical observation that in practice there is very little difference between choosing the leaf labels in either way. However, choosing the labels of pruned leaves according to the majority of pruning examples will set such leaves into a different status than the original leaves, which have as their label the majority class of training examples.

**Example** Figure 1 shows a decision tree that will be pruned into a single leaf if the training examples are used to label pruned leaves. A negative leaf replaces the root of the tree and makes two mistakes on the pruning examples, while the original tree makes three mistakes. With this tree we can illustrate an important difference in using training and pruning examples to label pruned leaves. Using training examples and proceeding bottom-up, observe that neither subtree is pruned, since the left one replaced with a negative leaf would make two mistakes instead of the original one mistake. Similarly, the right subtree replaced with a positive leaf would result in an increased number of classification errors. Nevertheless, the root node — even though its subtrees have not been pruned — can still be pruned.

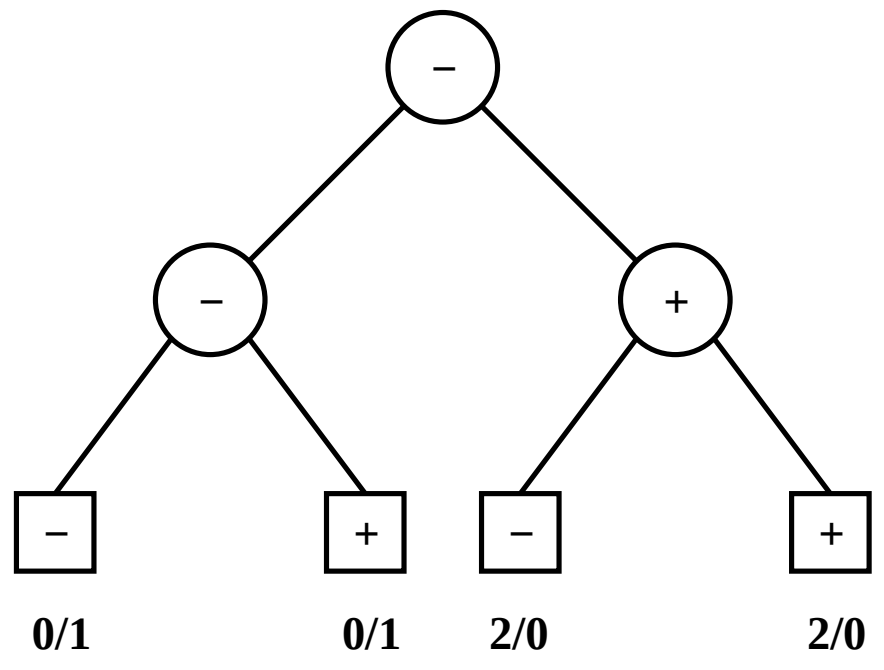

Figure 1: A (part of a) decision tree. The labels inside the nodes denote the majority classes of training examples arriving to these nodes. For leaves the numbers of pruning examples from the two classes are also given. $x/y$ means that $x$ negative and $y$ positive instances reach the leaf.

When pruning examples are used to label pruned leaves, a node with two non-trivial subtrees cannot be pruned unless both its subtrees are collapsed into leaves. The next section will prove this. In the tree of Figure 1 both subtrees would be collapsed into zero-error leaves. However, in this case the root node will not be pruned.

A further possibility for labeling the leaf nodes would be to take both training and pruning examples into account in deciding the label of a pruned leaf. Depending on the relation of the numbers of training and pruning examples this strategy resembles one or the other of the above-described approaches. Usually the training examples are more numerous than the pruning examples, and will thus dominate. In practice it is impossible to discern this labeling strategy from that of using the majority of training examples.

### 2.3 Empty Subtrees

Since REP uses different sets of examples to construct and to prune a decision tree, it is possible that some parts of the tree do not receive any examples in the pruning phase. Such parts of the decision tree, naturally, can be replaced with a single leaf without changing the number of classification errors that the tree makes on the pruning examples. In other words, subtrees that do not obtain any pruning examples are always pruned. Quinlan (1987) already noted that the parts of the original tree that correspond to rarer special cases, which are not represented in the pruning set, may be excised.

```
DecisionTree REP( DecisionTree T, ExampleArray S )
  { for ( i = 0 to S.length-1 ) classify( T, S[i] );
    return prune( T ); }

void classify( DecisionTree T, Example e )
  { T.total++; if ( e.label == 1 ) T.pos++;    // update node counters
    if ( !leaf(T) )
       if ( T.test(e) == 0 ) classify( T.left, e );
       else classify( T.right, e ); }

int prune( DecisionTree T )  // Output classification error after pruning T
  { if ( leaf(T) )
       if ( T.label == 1 ) return T.total - T.pos;
       else return T.pos;
    else
      { error = prune( T.left ) + prune( T.right );
        if ( error < min( T.pos, T.total - T.pos ) )
          return error;
        else
          { replace T with a leaf;
            if ( T.pos > T.total - T.pos )
              { T.label = 1; return T.total - T.pos; }
            else
              { T.label = 0; return T.pos; } } } }
```

Table 1: The REP algorithm. The algorithm first classifies the pruning examples in a top-down pass using method `classify` and then during a bottom-up pass prunes the tree using method `prune`.

Intuitively, it is not clear which is the best-founded strategy for handling *empty subtrees*, those that do not receive any examples. On one hand they obtain support from the training set, which usually is more numerous than the pruning set but, on the other hand, the fact that no pruning example corresponds to these parts of the tree would justify drawing the conclusion that these parts of the decision tree were built by chance properties of the training data. In REP, consistently with preferring smaller prunings also otherwise, the latter view is adopted.

The problem of empty subtrees is connected to the problem of *small disjuncts* in machine learning algorithms (Holte, Acker, & Porter, 1989). A small disjunct covers only a small number of the training examples. Collectively the small disjuncts are responsible for a small number of classification decisions, but they accumulate most of the error of the whole concept. Nevertheless, small disjuncts cannot be eliminated altogether, without adversely affecting other disjuncts in the concept.

### 2.4 The Analyzed Pruning Algorithm

Let us briefly reiterate the details of the REP algorithm that is analyzed subsequently. As already stated, the control strategy of the algorithm is the single-sweep bottom-up processing. First, a top-down traversal drives the pruning examples through the tree to the appropriate leaves. The counters of the nodes en route are updated. Second, during a bottom-up traversal the pruning operations indicated by the classification errors are executed. The errors can be determined on the basis of the node counter values. In the bottom-up traversal each node is visited only once. The pruned leaves are labeled by the majority of the pruning set (see Table 1).

## 3. Previous Work

Pruning of decision trees has recently received a lot of analytical attention; existing pruning methods have been analyzed (Esposito et al., 1993, 1997; Oates & Jensen, 1997, 1998, 1999) and new analytically-founded pruning techniques have been developed (Helmbold & Schapire, 1997; Pereira & Singer, 1999; Mansour, 1997; Kearns & Mansour, 1998). Also many empirical comparisons of pruning have appeared (Mingers, 1989a; Malerba, Esposito, & Semeraro, 1996; Frank, 2000). In this section we review earlier work that concerns the REP algorithm. Further related research is considered in Section 7.

Esposito et al. (1993) viewed the REP algorithm, among other pruning methods, as a search process in the state space. In addition to noting that the iterative version of REP cannot produce the optimal result required by Quinlan (1987), they also observed that even though REP is a linear-time algorithm in the size of the tree, with respect to the height of the tree REP requires exponential time in the worst case. In their subsequent comparative analysis Esposito et al. (1997) sketched a proof for Quinlan's (1987) claim that the pruning produced by REP is the smallest among the most accurate prunings of the given decision tree.

The bias of REP was briefly examined by Oates and Jensen (1997, 1998). They observed that the error, $r_L$, of the best majority leaf that could replace a subtree $T$ only depends on (the class distribution of) the examples that reach the root $N$ of $T$. In other words, the tree structure above $T$ and $N$ decides the error $r_L$. Let $r_T$ denote the error of the subtree $T$ at the moment when the pruning sweep reaches $N$; i.e., when some pruning may already have taken place in $T$. All pruning operations performed in $T$ have led either $r_T$ to decrease from the initial situation or to stay unchanged. In any case, pruning that has taken place in $T$ potentially decreases $r_T$, but does not affect $r_L$. Hence, the probability that $r_T < r_L$ — i.e., that $T$ will not be pruned — increases through pruning in $T$. This error propagation bias is inherent to REP. Oates and Jensen (1997, 1998) conjecture that the larger the original tree and the smaller the pruning set, the larger this effect, because a large tree provides more pruning opportunities and the high variance of a small pruning set offers more random chances for $r_L \leq r_T$. Subsequently we study some of these effects exactly.

In a follow-up study Oates and Jensen (1999) used REP as a vehicle for explaining the problems that have been observed in the pruning phase of top-down induction of decision trees. They analyzed REP in a situation in which the decision node under consideration fits noise — i.e., when the class of the examples is independent of the value of the attribute tested in the node at hand — and built a statistical model of REP in this situation. It indicates, consistently with their earlier considerations, that even though the probability of pruning a node that fits noise prior to pruning beneath it is close to 1, pruning that occurs beneath the node reduces its pruning probability close to 0. In particular, this model shows that if even one descendant of node $N$ at depth $d$ is not pruned, then $N$ will not be pruned (assuming there are no leaves until depth $d+1$). The consequence of this result is that increasing depth $d$ leads to an exponential decrease of the node's pruning probability.

The first part of Oates and Jensen's (1999) analysis is easy to comprehend, but its significance is uncertain, because this situation does not rise in any bottom-up pruning strategy. The statistical model is based on the assumption that the number, $n$, of pruning instances that pass through the node under consideration is large, in which case — independence assumptions prevailing — the errors committed by the node can be approximated by the normal distribution. The expected error of the original tree is the mean of the distribution, while, if pruned to a leaf, the tree would misclassify a proportion of the $n$ examples that corresponds to that of the minority class. Oates and Jensen show that the latter number is always less than the mean of the standard distribution of errors. Hence, the probability of pruning is over 0.5 and approaches 1 as $n$ grows.

In the second part of the analysis, in considering the pruning probability of a node $N$ after pruning has taken place beneath it, Oates and Jensen assume that the proportion of positive examples in any descendant of $N$ at depth $d$ is the same as in $N$. In this setting, assuming further that $N$ has a positive majority, all its descendants at level $d$ also have a positive majority. It directly follows that if all descendants at level $d$ are pruned, they are all replaced by a positive leaf. Hence, the function represented by this pruning is identically positive. The majority leaf that would replace $N$ also represents the same function and is smaller than the above pruning. Therefore, REP will choose the single leaf pruning. On the other hand, if one or more of the descendants of $N$ at depth $d$ are not pruned, then the pruning of the tree rooted at $N$, in which these subtrees are maintained and all other nodes at level $d$ are pruned into positive leaves, is more accurate than the majority leaf. In this case the tree will not be pruned.

Oates and Jensen (1999) also assume that starting from any node at level $d$ the probability of routing an example to a positive leaf is the same. In the following analyses we try to rid all unnecessary assumptions; the same results can be obtained without any knowledge of the example distribution.

## 4. Basic Properties of REP

Before going to the detailed probabilistic analysis of the REP algorithm, we examine some of its basic algorithmic properties. Throughout this paper we review the binary case for simplicity. The results, however, also apply with many-valued attributes and several classes.

Now that the processing control of the REP algorithm has been settled, we can actually prove Quinlan's (1987) claim of the optimality of the pruning produced by REP. Observe that the following result holds true independent of the leaf labeling strategy.

**Theorem 1** *Applying* REP *with a set of pruning examples, $S$, to a decision tree $T$ produces $T'$ — a pruning of $T$ — such that it is the smallest of those prunings of $T$ that have minimal error with respect to the example set $S$.*

**Proof** We prove the claim by induction over the size of the tree. Observe that a decision tree $T$ is a full binary tree, which has $2L(T) - 1$ nodes, where $L(T)$ is the number of leaves in the tree.

*Base case.* If $L(T) = 1$, then the original tree $T$ consists of a single leaf node. $T$ is the only possible pruning of itself. Thus, it is, trivially, also the smallest among the most accurate prunings of $T$.

*Inductive hypothesis.* The claim holds when $L(T) < k$.

*Inductive step.* $L(T) = k$. Let $N$ be the root of the tree and $T_0$ and $T_1$ the left and the right subtree, respectively. Subtrees $T_0$ and $T_1$ must have strictly less than $k$ leaves. When the pruning decision for $N$ is taken, then — by the bottom-up recursive control strategy of REP — $T_0$ and $T_1$ have already been processed by the algorithm. By the inductive hypothesis, the subtrees after pruning, $T'_0$ and $T'_1$, are the smallest possible among the most accurate prunings of these trees.

*(i): Accuracy.* The pruning decision for the node $N$ consists of choosing whether to collapse $N$ and the tree rooted at it into a majority leaf, or whether to maintain the whole tree. If both alternatives make the same number of errors, then $N$ is collapsed and the original accuracy with respect to the pruning set is retained. Otherwise, by the REP algorithm, the pruning decision is based on which of the resulting trees would make less errors with respect to the pruning set $S$. Hence, whichever choice is made, the resulting tree $T'$ will make the smaller number of errors with respect to $S$.

Let us now assume that a pruning $T''$ of $T$ makes even less errors with respect to $S$ than $T'$. Then $T''$ must consist of the root $N$ and two subtrees $T''_0$ and $T''_1$, because the majority leaf cannot be more accurate than $T'$. Since $T''$ is a more accurate pruning of $T$ than $T'$, it must be that either $T''_0$ is a more accurate pruning of $T_0$ than $T'_0$ or $T''_1$ is a more accurate pruning of $T_1$ than $T'_1$. By the inductive hypothesis both possibilities are false. Therefore, $T'$ is the most accurate pruning of $T$.

*(ii): Size.* To see that the chosen alternative is also as small as possible, first assume that $T'$ consists of a single leaf. Such a tree is the smallest pruning of $T$, and in this case the claim follows. Otherwise, $T'$ consists of the root node $N$ and the two pruned subtrees $T'_0$ and $T'_1$. Since this tree was not collapsed, the tree must be more accurate than the tree consisting of a single majority leaf. Now assume that there exists a pruning $T^*$ of $T$ that is as accurate as $T'$, but smaller. Because the majority leaf is less accurate than $T'$, $T^*$ must consist of the root node $N$ and two subtrees $T^*_0$ and $T^*_1$. Then, either $T^*_0$ is a smaller pruning of $T_0$ than $T'_0$, but as accurate, or $T^*_1$ is a smaller pruning of $T_1$ than $T'_1$, but as accurate. Both cases contradict the inductive hypothesis. Hence, $T'$ is the smallest among the most accurate prunings of $T$.

Thus, in any case, the claim follows for $T$. □

We consider next the situation in an internal node of the tree, when the bottom-up pruning sweep reaches the node. From now on we are committed to leaf labeling by the majority of the pruning examples.

**Theorem 2** *An internal node, which prior to pruning had no leaves as its children, will not be pruned by* REP *if it has a non-trivial subtree when the bottom-up pruning sweep reaches it.*

**Proof** For an internal node $N$ we have two possible cases in which it has non-trivial subtrees; either both its subtrees are non-trivial (non-leaf) or one of them is trivial. Let us review these cases.

Let $r_T$ denote the error of (sub)tree $T$ with respect to the part of the pruning set that reaches the root of $T$. By $r_L$ we denote the misclassification rate of the majority leaf $L$ that would replace $T$, if $T$ was chosen to be pruned.

Case I: Let the two subtrees of $T$, $T_0$ and $T_1$, be non-trivial. Hence, both of them have been retained when the pruning sweep has passed them. Thus, $r_{T_0} < r_{L_0}$ and $r_{T_1} < r_{L_1}$, where $L_0$ and $L_1$ are the majority leaves that would replace $T_0$ and $T_1$, respectively, if pruned. Because $r_T = r_{T_0} + r_{T_1}$, it must be that $r_T < r_{L_0} + r_{L_1}$.

If $T_0$ and $T_1$ have the same the majority class, then it is also the majority class of $T$. Then $r_L = r_{L_0} + r_{L_1}$, where $L$ is the majority leaf corresponding to $T$. Otherwise, $r_L \geq r_{L_0} + r_{L_1}$. In any case, $r_L \geq r_{L_0} + r_{L_1}$. Combining this with the fact that $r_T < r_{L_0} + r_{L_1}$ means that $r_T < r_L$. Hence, $T$ is not pruned.

Case II: Let $T$ have one trivial subtree, which was produced by pruning, and one non-trivial subtree. We assume, without loss of generality, that $T_0$ is non-trivial and $L_1$ is a majority leaf which has replaced $T_1$ in the pruning process. Then, $r_{T_0} < r_{L_0}$. Hence, we have that $r_T = r_{T_0} + r_{L_1} < r_{L_0} + r_{L_1}$.

In the same way as in the Case I, we can deduce that $r_L \geq r_{L_0} + r_{L_1}$. Therefore, $r_T < r_L$ and $T$ will be retained in the pruned tree.

$T$ cannot be pruned in either case, and the pruning process can be stopped on the branch containing $T$ unless an original leaf appears along the path from the root to $T$. □

If node $N$ has an original leaf, then it may be pruned even if the other subtree of $N$ is non-trivial. Also when $N$ has two trivial subtrees, it may be pruned. Whether pruning takes place depends on the class distribution of examples reaching $N$ and its subtrees.

In the analysis of Oates and Jensen (1999) it was shown that the prerequisite for pruning a node $N$ from the tree is that all its descendants at depth $d$ have been pruned. $d$ is the depth just above the first (original) leaf in the subtree rooted at $N$. If we apply the above result to this situation, we can corroborate their finding that $N$ will not be pruned if one or more of its descendants at depth $d$ are retained. Applying Theorem 2 recursively gives the result.

**Corollary 3** *A tree $T$ rooted at node $N$ will be retained by* REP *if one or more of the descendants of $N$ at depth $d$ are not pruned.*

To avoid the analysis being restricted by the leaf globally closest to the root, we need to be able to consider the set of leaves closest to the root on all branches of the tree. Let us define that the *fringe* of a decision tree contains any node that prior to pruning had a leaf as its child. Furthermore, any node that is in a subtree rooted at a node belonging to the fringe of the tree is also in the fringe. Those nodes not belonging to the fringe make up the *interior* of the tree. *Safe nodes* themselves belong to the fringe of the tree, but have their parent in the interior of the tree (see Figure 2). Because the fringe of a decision tree is closed downwards, the safe nodes of a tree correspond to the leaves of some pruning of it. Observe also that along the path from the root to a safe node there are no leaves. Therefore, if the pruning process ever reaches a safe node, Theorem 2 applies on the corresponding branch from there on.

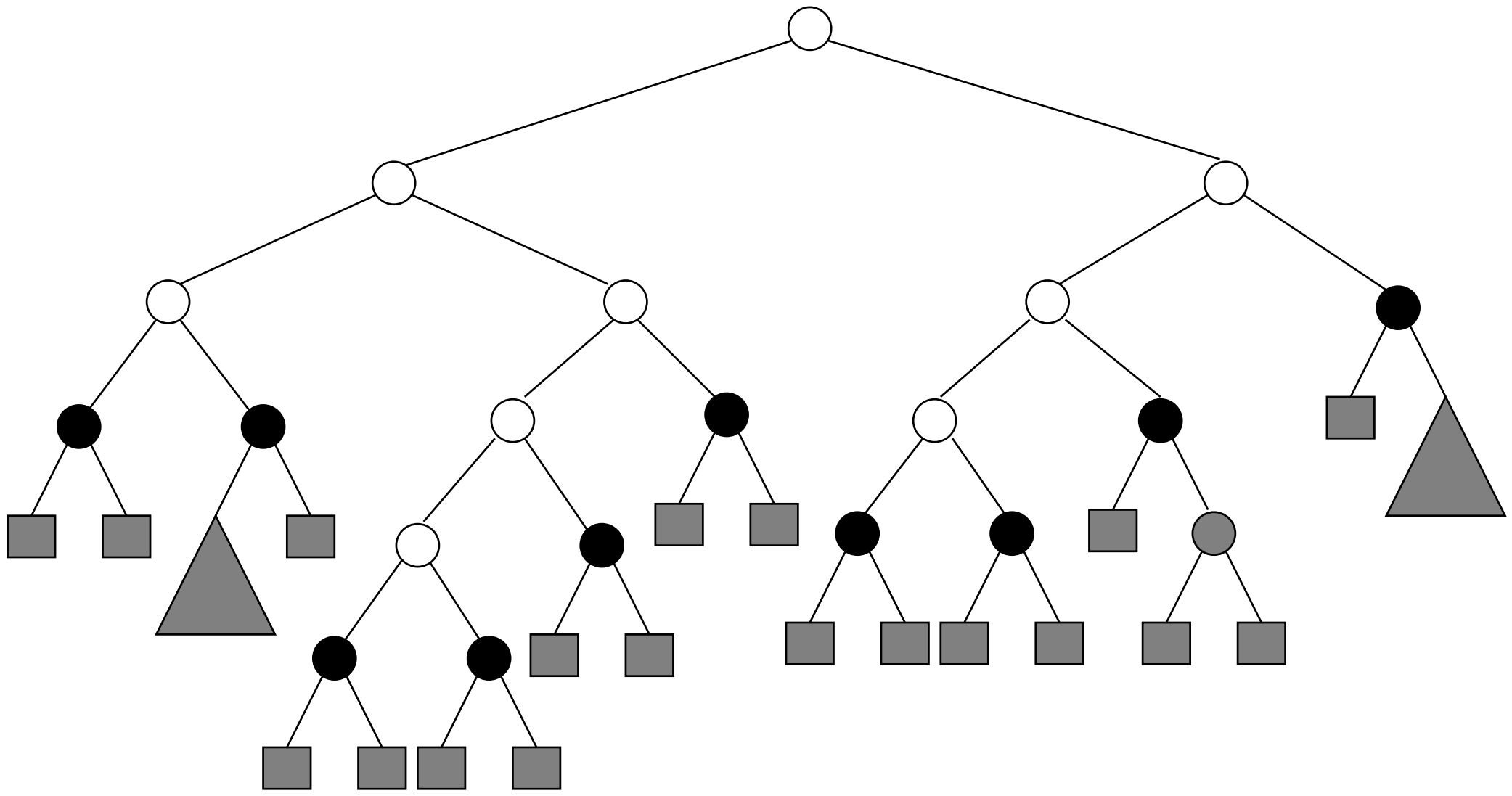

Figure 2: The fringe (black and gray nodes), interior (white nodes), and the safe nodes (black ones) of a decision tree. The triangles denote non-trivial subtrees.

If the decision tree under consideration will be pruned into a single majority leaf, safe nodes also need to be turned into leaves at some point, not necessarily simultaneously. If the pruning sweep continues to the safe nodes, from then on the question whether a node is pruned is settled solely on the basis of whether all nodes on the path to the root have the same majority class. The pruning of the whole tree can be characterized as below.

Let $T$ be the tree to be pruned and $S$ the set of pruning examples, $|S| = n$. We assume, without loss of generality, that at least half of the pruning examples are positive. Let $p$ be the proportion of positive examples in $S$; $p \geq 0.5$. If $T$ was to be replaced by a majority leaf, that leaf would have a positive class label. Under these assumptions we can prove the following.

**Theorem 4** *A tree $T$ will be pruned into a single leaf if and only if*

- *all subtrees rooted at the safe nodes of $T$ are pruned and*
- *at least as many positive as negative pruning examples reach each safe node in $T$.*

**Proof** To begin we show that the two conditions are necessary for the pruning of $T$. First, we show that if the former condition is not fulfilled, then $T$ cannot be pruned into a single leaf. Second, we prove that neither will $T$ be pruned if the former condition holds, but the latter not. Third, we show the sufficiency of the conditions; i.e., prove that if they both hold, then $T$ will be pruned into a single leaf.

(i): Let us first assume that in $T$ there is a safe node $N$ such that it will not be pruned. By the definition of a safe node, the parent $P$ of $N$ originally had no leaves as its children. Therefore, by Theorem 2, $P$ will not be pruned. It is easy to see, inductively, that neither will the root of $T$ be pruned.

(ii): Let us then assume that all subtrees rooted at safe nodes get pruned and that there are one or more safe nodes in $T$ into which more negative than positive pruning examples fall. Observe that all safe nodes cannot be such. Let us now consider the pruning of $T$ in which the leaves are situated in place of the safe nodes; the leaves receive the same examples as the original safe nodes. Because safe nodes are internal nodes, in REP the corresponding pruned leaves are labeled by the majority of the pruning examples. In particular, the safe nodes that receive more negative than positive examples are replaced by negative leaves. All other leaves are labeled positive. This pruning of the original tree is more accurate than the majority leaf. Hence, by Theorem 1, REP will not prune $T$ into a single-leaf tree.

(iii): Let us now assume that all subtrees rooted at the safe nodes of $T$ are pruned and that at least as many positive as negative pruning examples reach each safe node. Then all interior nodes must also have a majority of positive pruning examples. Otherwise, there is an interior node $N$ in $T$ that has more negative than positive examples. Thus, at least one of the children of $N$ has a majority of negative examples. Carrying the induction all the way to the safe nodes shows that no such node $N$ can exist in $T$. Hence, all interior prunings of $T$ represent the same function (identically positive) and all of them have the same error with respect to $S$. The majority leaf is the unique, smallest of these prunings and will, by Theorem 1, be chosen.

□

## 5. A Probabilistic Analysis of REP

Let us now turn our attention to the question of what the prerequisites for pruning a decision tree $T$ into a single majority leaf are. Since, by Theorem 1, REP produces a pruning of $T$ which is the most accurate with respect to the pruning set and such that it is as small as possible, to show that $T$ does not reduce to a single leaf it suffices to find its pruning that has a better prediction accuracy on the pruning examples than the majority leaf has.

In the following the class of an example is assumed to be independent of its attribute values. Obviously, if in a decision tree there is a node where this assumption holds for the examples arriving to it, we would like the pruning algorithm to turn it into a majority leaf. We do not make any assumptions about the decision tree. However, similar to the analysis of Oates and Jensen (1999), for the obtained bounds to be tight, the shortest path from the root of the tree to a leaf should not be too short.

### 5.1 Probability Theoretical Preliminaries

Let us recall some basic probabilistic concepts and results that are used subsequently. We denote the probability of an event $E$ by $\mathbf{Pr}\{E\}$ and its expectation by $\mathbf{E}E$. A discrete (integer-valued) random variable $X$ is said to be *binomially distributed with parameters $n$ and $p$*, denoted by $X \sim B(n,p)$, if

$$\mathbf{Pr}\{\, X = k \,\} = \binom{n}{k} p^k (1-p)^{n-k}, \quad k = 0, 1, \ldots, n.$$

If $X \sim B(n,p)$, then its expected value or mean is $\mathbf{E}X = \mu = np$, variance $\mathbf{var}X = np(1-p)$, and standard deviation $\sigma = \sqrt{np(1-p)}$.

An *indicator variable* is is a discrete random variable that takes on only the values 0 and 1. An indicator variable $I$ is used to denote the occurrence or non-occurrence of an event. If $A_1, \ldots, A_n$ are independent events with $\mathbf{Pr}\{A_i\} = p$ and $I_{A_1}, \ldots, I_{A_n}$ are the respective indicator variables, then $X = \sum_{i=1}^{n} I_{A_i}$ is binomially distributed with parameters $n$ and $p$. $I_{A_i}$ is called a *Bernoulli* random variable with parameter $p$.

The *density function* $f_X : \mathbb{N} \to [0,1]$ for a discrete random variable $X$ is defined as $f_X(x) = \mathbf{Pr}\{\, X = x \,\}$. The *cumulative distribution function* $F_X : \mathbb{N} \to [0,1]$ for $X$ is defined as $F_X(y) = \mathbf{Pr}\{\, X \le y \,\} = \sum_{x \le y} f_X(x)$.

Let $X \sim B(n,p)$ be a random variable with mean $\mu = np$ and standard deviation $\sigma = \sqrt{np(1-p)}$. The *normalized random variable* corresponding to $X$ is

$$\widetilde{X} = \frac{X - \mu}{\sigma}.$$

By the central limit theorem we can approximate the cumulative distribution function $F_{\widetilde{X}}$ of $\widetilde{X}$ by the *normal* or *Gaussian distribution*

$$F_{\widetilde{X}}(y) = \mathbf{Pr}\left\{ \widetilde{X} \le y \right\} \approx \Phi(y).$$

$\Phi$ is the cumulative distribution function of the “bell curve” density function $e^{-x^2/2}/\sqrt{2\pi}$.

Respectively, we can apply the *normal approximation* to the corresponding random variable $X$

$$F_X(y) = \mathbf{Pr}\{\, X \le y \,\} = F_{\widetilde{X}}\left(\frac{y-\mu}{\sigma}\right) \approx \Phi\left(\frac{y-\mu}{\sigma}\right).$$

### 5.2 Bounding the Pruning Probability of a Tree

Now, the pruning set is considered to be a sample from a distribution in which the class attribute is independent of the other attributes. We assume that the class attribute is distributed according to Bernoulli($p$) distribution; i.e., the class is positive with probability $p$ and negative with probability $1-p$. We assume that $p > 0.5$.

In the following we will analyze the situation in which the subtrees rooted at safe nodes have already been pruned into leaves. We bound the pruning probability of the tree starting from this initial configuration. Since the bottom-up pruning may already have come to a halt before that situation, the following results actually give too high a probability for pruning. Hence, the following upper bounds are not as tight as possible.

We consider pruning a decision tree by REP as a trial whose result is decided by the set of pruning examples. By Theorem 4 we can approximate the probability that a tree will be pruned into a majority leaf by approximating the probability that *all* safe nodes get a positive majority or a negative majority. The latter alternative is not very probable under the assumption $p > .5$. It is safe to assume that it never happens.

We can consider sampling the pruning examples in two phases. First the attribute values are assigned. This decides the leaf into which the example falls. In the second phase we independently assign the class label for the example.

Let the safe nodes of tree $T$ be $Z(T) = \{z_1, \ldots, z_k\}$ and let the number of examples in the pruning set $S$ be $|S| = n$. The number of pruning examples falling to a safe node $z_i$ is denoted by $n_i$; $\sum_{i=1}^{k} n_i = n$. For the time being we assume that $n_i > 0$ for all $i$. The number of positive examples falling to safe node $z_i$ is the sum of independent Bernoulli variables and, thus, it is binomially distributed with parameters $n_i$ and $p$. Respectively, the number of negative pruning examples in safe node $z_i$ is $X_i \sim B(n_i, 1-p)$. The probability that there is a majority of negative examples in safe node $z_i$ is $\mathbf{Pr}\{\, X_i > n_i/2 \,\}$. We can bound this probability from below by using the following inequality (Slud, 1977).

**Lemma 5 (Slud's inequality)** *Let $X \sim B(m, q)$ be a random variable with $q \leq 1/2$. Then for $m(1-q) \geq h \geq mq$,*

$$\mathbf{Pr}\{\, X \geq h \,\} \geq 1 - \Phi\left(\frac{h - mq}{\sqrt{mq(1-q)}}\right).$$

Since $p > .5$ and the random variable corresponding to the number of negative examples in safe node $z_i$ is $X_i \sim B(n_i, 1-p)$, the first condition of Slud's inequality holds. Furthermore, to see that condition $m(1-q) \geq h \geq mq$ holds in safe node $z_i$ substitute $h = n_i/2$, $m = n_i$, and $q = 1 - p$ to obtain $n_i p \geq n_i/2 \geq n_i(1-p)$. Thus,

$$\mathbf{Pr}\left\{ X_i > \frac{n_i}{2} \right\} \geq 1 - \Phi\left(\frac{n_i/2 - n_i(1-p)}{\sqrt{n_i p(1-p)}}\right) = 1 - \Phi\left(\frac{(p - 1/2)n_i}{\sqrt{n_i p(1-p)}}\right). \qquad (1)$$

As $n_i$, the number of pruning instances reaching safe node $z_i$, grows, then the standard normal distribution term in the above bound also grows. Hence, the bound on the probability that the majority of the pruning examples reaching $z_i$ is negative is the smaller the more pruning examples reach it. The probability of a negative majority also reduces through the growing probability of positive class for an example, $p$. These both are also reflected in the pruning probabilities of the whole tree.

We can now roughly approximate the probability that $T$ will be pruned into a single majority leaf as follows. By Theorem 4, $T$ will be pruned into a leaf if and only if each safe node in $T$ receives a majority of positive examples. Because $T$ has $k$ safe nodes and there are $n$ pruning examples, then according to the pigeon-hole principle at least half of the safe nodes receive at most $r = 2n/k$ examples. Each safe node $z_i$ with $n_i \leq r$ examples has, by Inequality 1, a negative majority at least with probability

$$1 - \Phi\left(\frac{(p - 1/2)r}{\sqrt{rp(1-p)}}\right).$$

Observe that Inequality 1 also holds when $n_i < r$, because the cumulative distribution function $\Phi$ is an increasing function. The argument $n_i(p-1/2)/\sqrt{n_i p(1-p)}$ can be rewritten as $\sqrt{n_i}c_p$, where $c_p$ is a positive constant depending on the value of $p$. Since $\Phi(\sqrt{n_i}c_p)$ grows as $n_i$ grows, $1 - \Phi(\sqrt{n_i}c_p)$ grows with decreasing $n_i$. Hence, the lower bound of Inequality 1 also applies for values $0 < n_i < r$.

Thus, the probability that the half of the safe nodes that receive at most $r$ examples have a positive majority is at most

$$\left(\Phi\left(\frac{(p-1/2)r}{\sqrt{rp(1-p)}}\right)\right)^{k/2}. \tag{2}$$

This is an upper bound for the probability that the whole tree $T$ will be pruned into a single leaf. The only distribution assumption that was made to reach the result is that $p > .5$. In order to obtain tighter bounds, one has to make assumptions about the shape of the tree $T$ and the distribution of examples.

The bound of Equation 2 depends on the size of the decision tree (reflected by $k$), the number ($n$) and the class distribution ($p$) of the pruning examples. Keeping other parameters constant and letting $k$ grow reduces the pruning probability exponentially. If the number of pruning examples grows in the same proportion so that $r = 2n/k$ stays constant, the pruning probability still falls exponentially. Class distribution of the pruning examples also affects the pruning probability which is the smaller, the closer $p$ is to value .5.

### 5.3 Implications of the Analysis

It has been empirically observed that the size of the decision tree grows linearly with the training set size, even when the trees are pruned (Catlett, 1991; Oates & Jensen, 1997, 1998). The above analysis gives us a possibility to explain this behavior. However, let us first prove that when there is no correlation between the attribute values and the class label of an example, the size of the tree that perfectly fits the training data depends linearly on the size of the sample.

Our setting is as simple as can be. We only have one real-valued attribute $x$ and the class attribute $y$, whose value is independent of that of $x$. As before, $y$ has two possible values, 0 and 1. The tree is built using binary splits of a numerical value range; i.e., propositions of type "$x < r$" are assigned to the internal leaves of the tree. In this analysis duplicate instances occur with probability 0.

**Theorem 6** *Let the training examples $(x, y)$ be drawn from a distribution, where $x$ is uniformly distributed in the range $[0, 1)$ and $y$ obtains value 1, independent of $x$, with probability $p$, and value 0 with probability $1 - p$. Then the expected size of the decision tree that fits the data is linear in the size of the sample.*

**Proof** Let $S = \langle (x_1, y_1), \ldots, (x_t, y_t) \rangle$ be a sample of the above described distribution. We may assume that $x_i \neq x_j$, when $i \neq j$, because the probability of the complement event is 0. Let us, further, assume that the examples of $S$ have been indexed so that $x_1 < x_2 < \ldots < x_t$. Let $A_i$ be the indicator variable for the event that instances $i$ and $i+1$ have different class labels; i.e., $y_i \neq y_{i+1}$, $1 \leq i \leq t-1$. Then $\mathbf{E}A_i = \mathbf{Pr}\{\, A_i = 1 \,\} = p(1-p) + (1-p)p = 2p(1-p)$, because when the event $y_i = 1$ has probability $p$, at the same time the event $y_{i+1} = 0$ has probability $1 - p$, and vice versa. Now the number of class alternations is $A = \sum_{i=1}^{t-1} A_i$ and its expectation is

$$\mathbf{E}A = \sum_{i=1}^{t-1} \mathbf{E}A_i = \sum_{i=1}^{t-1} 2p(1-p) = 2p(1-p) \sum_{i=1}^{t-1} 1 = 2(t-1)p(1-p). \tag{3}$$

Let $T$ be a decision tree that has been grown on the sample $S$. The growing has been continued until the training error is 0. Each leaf in $T$ corresponds to a half open interval $[a, b)$ in $[0, 1)$. If $y_i \neq y_{y+1}$, then $x_i$ and $x_{i+1}$ must fall into different leaves of $T$, because otherwise one or the other example is falsely classified by $T$. Thus, the upper boundary $b$ of the interval corresponding to the leaf into which $x_i$ falls in must have a value less than $x_{i+1}$. Repetitively applying this observation when scanning through the examples from left to right, we see that $T$ must at least have one leaf for $x_1$ and one leaf for each class alternation; i.e., $A + 1$ leaves in total. By using Equation 3 we see that the expected number of leaves in $T$ is

$$\mathbf{E}A + 1 = 2(t-1)p(1-p) + 1.$$

In particular, this is linear in the size of the sample $S$; $|S| = t$. □

The above theorem only concerns zero training error trees built in the first phase of decision tree induction. The empirical observations of Catlett (1991) and Oates and Jensen (1997, 1998), however, concern decision trees that have been pruned in the second phase of induction. We come back to the topic of pruned trees shortly.

Consider how REP is used in practice. There is some amount of (classified) data available from the application domain. Let there be a total of $t$ examples available. Some part $\alpha$ of the data is used for tree growing and the remaining portion $1 - \alpha$ of it is reserved as the separate pruning set; $0 < \alpha < 1$. Quite a common practice is to use two thirds of the data for growing and one third for pruning or nine tenths for growing and one tenth for pruning when (ten-fold) cross-validation is used. In the decision tree construction phase the tree is fitted to the $\alpha t$ examples as perfectly as possible. If we hypothesize that the previous result holds for noisy real-world data sets, which by empirical evidence would appear to be the case, and that the number of safe nodes also grows linearly with the number of leaves, then the tree grown will contain $\gamma t$ safe nodes, where $\gamma > 0$. Since the pruning set size also is a linear fraction of the training set size, the ratio $r = 2n/k$ stays constant in this setting. Hence, by Equation 2, the growing data set size forces the pruning probability to zero, even quite fast, because the reduction in the probability is exponential.

### 5.4 Limitations of the Analysis

Empty subtrees, which do not receive any pruning examples, were left without attention above; we assumed that $n_i > 0$ for each $i$. Empty subtrees, however, decisively affect the analysis; they are automatically pruned away. Unfortunately, one cannot derive a non-trivial upper bound for the number of empty subtrees. In the worst case all pruning examples are routed to the same safe node, which leaves $k - 1$ empty safe nodes to the tree. Subsequently we review the case where the examples are distributed uniformly to the safe nodes. Then better approximations can be obtained.

Even though we assume that each pruning example is positive with a higher probability than .5, there are no guarantees that the majority of all examples is positive. However, the probability that the majority of all examples changes is very small, even negligible, by Chernoff's inequality (Chernoff, 1952; Hagerup & Rüb, 1990) when the number of pruning examples, $n$, is high and $p$ is not extremely close to one half.

Slud's inequality bounds the probability $\mathbf{Pr}\{ X \geq h \}$, but above we used it to bound the probability $\mathbf{Pr}\{ X > h \}$. Some continuity correction could be used to compensate this. In practice, the inexactness does not make any difference.

Even though it would appear that the number of safe nodes increases in the same proportion as that of leaves when the size of the training set grows, we have not proved this result. Theorem 6 essentially uses leaf nodes, and does not lend itself to modification, where safe nodes could be substituted in place of leaves.

The relation between the number of safe nodes and leaves in a decision tree depends on the shape of the tree. Hence, the splitting criterion that was used in tree growing decisively affects this relation. Some splitting criteria aim at keeping the produced split as balanced as possible, while others aim at separating small class coherent subsets from the data (Quinlan, 1986; Mingers, 1989b). For example, the common entropy-based criteria have a bias that favors balanced splits (Breiman, 1996). Using a balanced splitting criterion would seem to imply that the number of safe nodes in a tree depends linearly on the number of leaves in the tree. In that case the above reasoning would explain the empirically observed linear growth of pruned decision trees.

## 6. Pruning Probability Under Uniform Distribution

We now assume that all $n$ pruning examples have an equal probability to end up in each of the $k$ safe nodes; i.e., a pruning example falls to the safe node $z_i$ with probability $1/k$. Contrary to the normal uniform distribution assumption analysis, for our analysis this is not the best case. Here the best distribution of examples into safe nodes would have one pruning example in each of the safe nodes except one, into which all remaining pruning instances would gather. Nevertheless, the uniformity lets us sharpen the general approximation by using standard techniques.

The expected number of examples falling into any safe node is $n/k$. Let us calculate the expected number of those safe nodes that receive at most $cn/k$ examples, where $c$ is an arbitrary positive constant. Let $Q_i$ be the indicator for the event "safe node $z_i$ receives at most $cn/k$ examples." Then $Q = \sum_{i=1}^{k} Q_i$ is the number of those safe nodes that receive less than $cn/k$ examples. By the linearity of expectation $\mathbf{E}Q = \sum_{i=1}^{k} \mathbf{E}Q_i = k\mathbf{E}Q_1$, in which the last equality follows from the fact that the $Q_i$-s are identically distributed.

Let $Y_1$ be the number of examples reaching safe node $z_1$. Because each of the $n$ examples reaches $z_1$ with probability $1/k$ independent of the other examples, $Y_1$ is binomially distributed with parameters $n$ and $1/k$. Clearly $\mathbf{E}Q_1 = \mathbf{Pr}\{ Y_1 \leq cn/k \}$. We can approximate the last probability by the normal approximation, from which we obtain

$$\mathbf{Pr}\left\{ Y_1 \leq \frac{cn}{k} \right\} \approx \Phi\left( \frac{cn/k - n/k}{\sqrt{n \cdot 1/k \cdot (1 - 1/k)}} \right) = \Phi\left( \frac{(c-1)n/k}{\sqrt{n/k(1 - 1/k)}} \right).$$

Hence, by the above observation,

$$\mathbf{E}Q = k\mathbf{E}Q_1 \approx k\Phi\left(\frac{(c-1)n/k}{\sqrt{n/k(1-1/k)}}\right). \tag{4}$$

We now use Approximation 4 to determine the probability that the whole decision tree $T$ will be pruned into a single leaf. Let $P$ be a random variable that represent the number of those safe nodes in $T$ that receive at most $cn/k$ examples and at least one example. If we denote by $R$ the number of empty safe nodes, we have $P = Q - R$. Hence, $\mathbf{E}P = \mathbf{E}(Q-R) = \mathbf{E}Q - \mathbf{E}R$.

The following result (Kamath, Motwani, Palem, & Spirakis, 1994; Motwani & Raghavan, 1995) lets us approximate the number of empty safe nodes when $n \gg k$.

**Theorem 7** *Let $Z$ be the number of empty bins when $m$ balls are thrown randomly into $h$ bins. Then*

$$\mu = \mathbf{E}Z = h\left(1-\frac{1}{h}\right)^m \approx he^{-m/h}$$

*and for $\lambda > 0$,*

$$\mathbf{Pr}\{\,|Z-\mu| \geq \lambda\,\} \leq 2\exp\left(-\frac{\lambda^2(h-1/2)}{h^2-\mu^2}\right).$$

By this result the expected number of empty safe nodes is approximately $ke^{-n/k}$; this number is small when $k$ is relatively small compared to $n$.

Substituting the above obtained approximation for $\mathbf{E}Q$ (Equation 4) and using the previous result, we get

$$\mathbf{E}P = \mathbf{E}Q - \mathbf{E}R \approx k\left(\Phi\left(\frac{(c-1)n/k}{\sqrt{n/k(1-1/k)}}\right) - e^{-n/k}\right).$$

Applying Slud's inequality we can, as before, bound from above the probability that the majority class does not change in a safe node that receives $cn/k$ pruning examples. Since there are $P$ such safe nodes and the class distribution of examples within them is independent, the event "majority class does not change in any safe node that receives at least one and at most $cn/k$ examples" has the upper bound

$$\left(\Phi\left(\frac{(p-.5)r}{\sqrt{rp(1-p)}}\right)\right)^P, \tag{5}$$

where $r = cn/k$. Replacing $P$ with its expected value in this equation we have an approximation for the pruning probability. This approximation is valid if $P$ does not deviate a lot from its expected value. We consider the deviation of $P$ from its expected value below.

The above upper bound for the pruning probability is similar to the upper bound that was obtained without any assumptions about the distribution of the examples. However, the earlier constant 2 has been replaced by a new, controllable parameter $c$, and empty subtrees are now explicitly taken into account. If $c$ is chosen suitably, this upper bound is more strict than the one obtained in the general case.

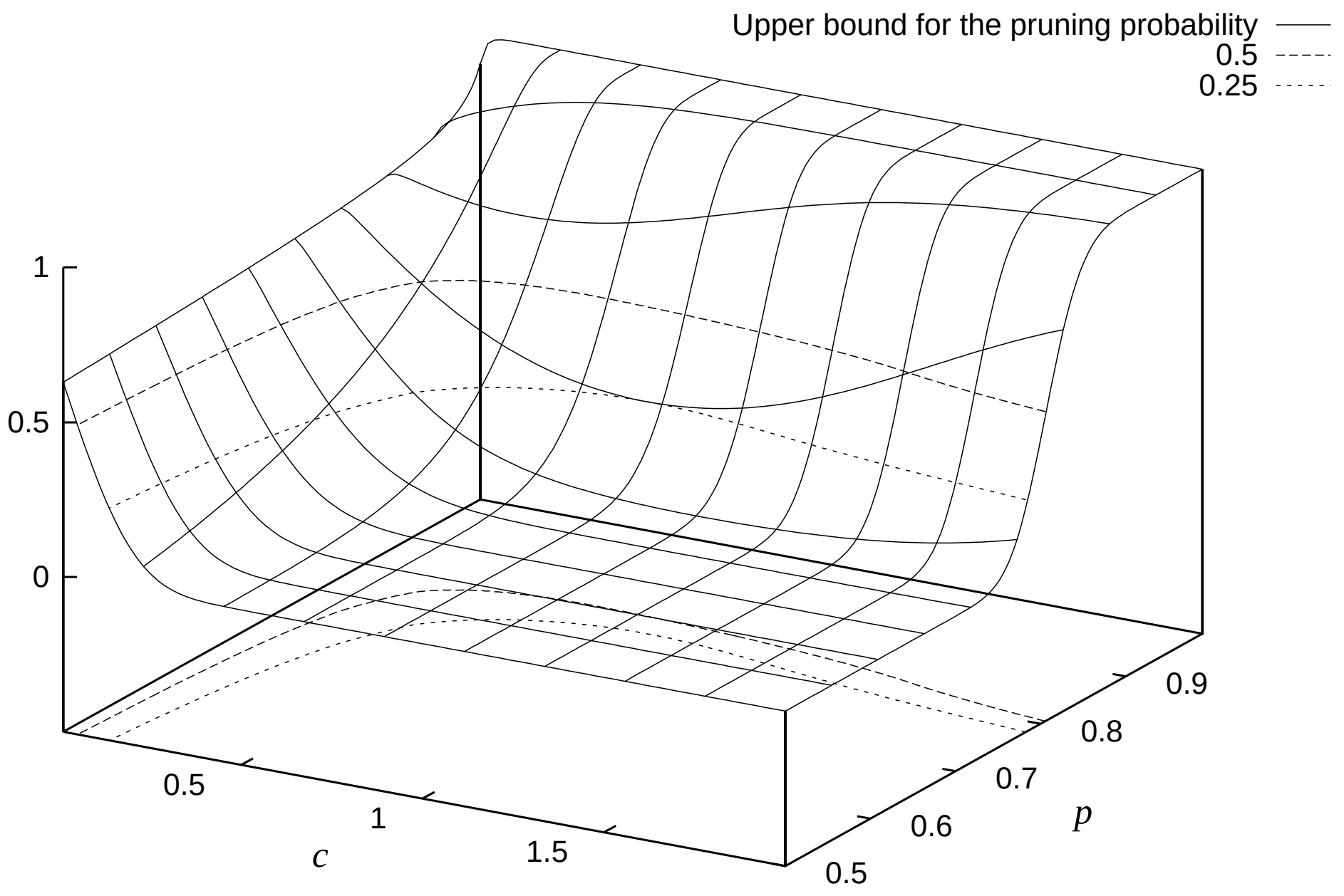

Figure 3: The effect of parameters $p$ and $c$ on the upper bound of the pruning probability of a tree with 100 safe nodes when 500 pruning examples are used. The curves depicting the 0.25 and 0.5 upper bounds are also shown.

### 6.1 An Illustration of the Upper Bound

Figure 3 plots the upper bound of the pruning probability of a tree with 100 safe nodes when 500 pruning examples are used. The value of the parameter $c$ varies from 0 to 2 and $p$ varies from 0.5 to 1. We can observe that the surface corresponding to the upper bound stays very close to 0 when the class distribution is not too skewed and when the parameter $c$ does not have a very small value. When the probability of an example having a positive class label hits value 0.75 or the value of $c$ approaches 0, the upper bound climbs very steeply. At least on the part of the parameter $c$ this is due to the inexactness of the approximation on the extreme values.

When the probability $p$ that an example has a positive class approaches 1, the error committed by a single positive leaf falls to 0. Hence, the accuracy of a non-trivial pruning has to be better, the closer $p$ is to 1 for it to beat the majority leaf. Intuitively, the probability that such a pruning exists — i.e., that the root node is not pruned — should drop to zero as $p$ increases. The bound reflects this intuition.

When the value of parameter $c$ falls close to 0, the safe nodes that are taken into account in the upper bound only receive very few pruning examples. The number of such nodes is small. On the other hand, when $c$ is increased, the number of nodes under consideration grows together with the upper limit on the number of examples reaching each single one of them. Thus, both small and large values of $c$ yield loose bounds. In the strictest bounds the value of $c$ is somewhere in "the middle," in our example around values 1.0–1.5. In the bound of Equation 5 the argument of the cumulative distribution function $\Phi$ tends towards zero when the value of $c$ is very small, but at the same time the exponent decreases. The value of $\Phi$ approaches $1/2$, when its argument goes to zero. On the other hand, when $c$ has a large value, $\Phi$ approaches value 1 and the exponent $P$ also increases.

### 6.2 On the Exactness of the Approximation

Above we used the expected value of $P$ in the analysis; $\mathbf{E}P = \mathbf{E}Q - \mathbf{E}R$. We now probe into the deviation of $P$ from its expected value. The deviation of $R$ is directly available from Theorem 7:

$$\mathbf{Pr}\{\, |R - \mathbf{E}R| \geq \lambda \,\} \leq 2\exp\left(-\frac{\lambda^2(k-1/2)}{k^2 - \mathbf{E}^2R}\right).$$

For $Q$ we do not have a similar result yet. In this section we provide one.

Let us first recapitulate the definition of the *Lipschitz condition.*

**Definition** Let $f : \mathcal{D}_1 \times \cdots \times \mathcal{D}_m \to \mathbb{R}$ be a real-valued function with $m$ arguments from possibly distinct domains. The function $f$ is said to satisfy the Lipchitz condition if for any $x_1 \in \mathcal{D}_1, \ldots, x_m \in \mathcal{D}_m$, any $i \in \{1, \ldots, m\}$, and any $y_i \in \mathcal{D}_i$,

$$|f(x_1, \ldots, x_{i-1}, x_i, x_{i+1}, \ldots, x_m) - f(x_1, \ldots, x_{i-1}, y_i, x_{i+1}, \ldots, x_m)| \leq 1.$$

Hence, a function satisfies the Lipschitz condition if an arbitrary change in the value of any one argument does not change the value of the function more than 1.

The following result (McDiarmid, 1989) holds for functions satisfying the Lipschitz condition. More general results of the same kind can be obtained using *martingales* (see e.g., (Motwani & Raghavan, 1995)).

**Theorem 8 (McDiarmid)** *Let $X_1, \ldots, X_m$ be independent random variables taking values in a set $V$. Let $f : V^m \to \mathbb{R}$ be such that, for $i = 1, \ldots, m$:*

$$\sup_{x_1,\ldots,x_m,y_i \in V} |f(x_1, \ldots, x_{i-1}, x_i, x_{i+1}, \ldots, x_m) - f(x_1, \ldots, x_{i-1}, y_i, x_{i+1}, \ldots, x_m)| \leq c_i.$$

*Then for $\lambda > 0$,*

$$\mathbf{Pr}\{\, |f(X_1, \ldots, X_m) - \mathbf{E}f(X_1, \ldots, X_m)| \geq \lambda \,\} \leq 2\exp\left(-\frac{2\lambda^2}{\sum_{i=1}^m c_i^2}\right).$$

Let $W_i$, $i = 1, \ldots, n$, be a random variable such that $W_i = j$ if the $i$-th example is directed to the safe node $z_j$. By the uniform distribution assumption $W_i$-s are independent. They have their values within the set $\{1, \ldots, k\}$. Let us define the function $f$ so that $f(w_1, \ldots, w_n)$ is the number of those safe nodes that receive at most $r = cn/k$ examples, when the $i$-th example is directed to the safe node $z_{w_i}$. That is,

$$f(w_1, \ldots, w_n) = |\{\, i \in \{\, 1, \ldots, k \,\} \mid |S_i| \leq r \,\}|,$$

where $S_i$ is the set of those examples that are directed to safe node $z_i$;

$$S_i = \{\, h \in \{\,1, \ldots, n\,\} \mid w_h = i\,\}.$$

Hence, $Q = f(W_1, \ldots, W_n)$. Moving any one example from one safe node to another (changing the value of any one argument $w_i$), can change one more safe node $z_i$ to fulfill the condition $|S_i| \leq r$, one less safe node to fulfill it, or both at the same time. Thus, the value of $f$ changes by at most 1. Hence, the function fulfills the Lipschitz condition. Therefore, we can apply McDiarmid's inequality to it by substituting $c_i = 1$ and observing that then $\sum_{i=1}^{n} c_i^2 = n$:

$$\mathbf{Pr}\{\, |f(W_1, \ldots, W_n) - \mathbf{E}f(W_1, \ldots, W_n)| \geq \lambda\,\} \leq 2e^{-2\lambda^2/n},$$

or equally

$$\mathbf{Pr}\{\, |Q - \mathbf{E}Q| \geq \lambda\,\} \leq 2e^{-2\lambda^2/n}.$$

Unfortunately, this concentration bound is not very tight. Nevertheless, combining the concentration bounds for $Q$ and $R$ we have for $P$ the following deviation from its expected value.

Since $|P - \mathbf{E}P| = |Q - R - \mathbf{E}(Q - R)| = |Q - \mathbf{E}Q + \mathbf{E}R - R| \leq |Q - \mathbf{E}Q| + |R - \mathbf{E}R|$, $|Q - R - \mathbf{E}(Q - R)| \geq \lambda$ implies that $|Q - \mathbf{E}Q| \geq \lambda/2$ or $|R - \mathbf{E}R| \geq \lambda/2$. Thus,

$$\begin{aligned}
\mathbf{Pr}\{\, |P - \mathbf{E}P| \geq \lambda\,\} &= \mathbf{Pr}\{\, |Q - R - \mathbf{E}(Q - R)| \geq \lambda\,\} \\
&\leq \mathbf{Pr}\left\{ |Q - \mathbf{E}Q| \geq \frac{\lambda}{2} \right\} + \mathbf{Pr}\left\{ |R - \mathbf{E}R| \geq \frac{\lambda}{2} \right\} \\
&\leq 2\exp\left( -\frac{\lambda^2}{2n} \right) + 2\exp\left( -\frac{\lambda^2(k - 1/2)}{4(k^2 - \mathbf{E}^2 R)} \right).
\end{aligned}$$

## 7. Related Work

Traditional pruning algorithms — like cost-complexity pruning (Breiman et al., 1984), pessimistic pruning (Quinlan, 1987), minimum error pruning (Niblett & Bratko, 1986; Cestnik & Bratko, 1991), critical value pruning (Mingers, 1989a), and error-based pruning (Quinlan, 1993) — have already been covered extensively in earlier work (Mingers, 1989a; Esposito et al., 1997; Frank, 2000). Thus we will not touch on these methods any further. Instead, we review some of the more recent work on pruning.

REP produces an optimal pruning of the given decision tree with respect to the pruning set. Other approaches for producing optimal prunings have also been presented (Breiman et al., 1984; Bohanec & Bratko, 1994; Oliver & Hand, 1995; Almuallim, 1996). However, often optimality is measured over the training set. Then it is only possible to maintain the initial accuracy, assuming that no noise is present. Neither is it usually possible to reduce the size of the decision tree without sacrificing the classification accuracy. For example, in the work of Bohanec and Bratko (1994) it was studied how to efficiently find the optimal pruning in the sense that the output decision tree is the smallest pruning which satisfies a given accuracy requirement. A somewhat improved algorithm for the same problem was presented subsequently by Almuallim (1996).

The high level control of Kearns and Mansour's (1998) pruning algorithm is the same bottom-up sweep as in REP. However, the pruning criterion in their method is a kind of a *cost-complexity* condition (Breiman et al., 1984) that takes both the observed classification error and (sub)tree complexity into account. Moreover, their pruning scheme does not require the pruning set to be separate from the training set. Both Mansour's (1997) and Kearns and Mansour's (1998) algorithms are *pessimistic*: they try to bound the true error of a (sub)tree by its training error. Since the training error is by nature optimistic, the pruning criterion has to compensate it by being pessimistic about the error approximation.

Consider yet another variant of REP, one which is otherwise similar to the one analyzed above, with the exception that the original leaves are not put to a special status, but can be relabeled by the majority of the pruning examples just like internal nodes. This version of REP produces the optimal pruning with respect to which the performance of Kearns and Mansour's (1998) algorithm is measured. Their pessimistic pruning produces a decision tree that is smaller than that produced by REP.

Kearns and Mansour (1998) are able to prove that their algorithm has a strong performance guarantee. The generalization error of the produced pruning is bounded by that of the best pruning of the given tree plus a complexity penalty. The pruning decisions are local in the same sense as those of REP and only the basic pruning operation of replacing a subtree with a leaf is used in this pruning algorithm.

## 8. Conclusion

In this paper the REP algorithm has been analyzed in three different settings. First, we studied the algorithmic properties of REP alone, without assuming anything about the input decision tree nor pruning set. In this setting it is possible to prove that REP fulfills its intended task and produces an optimal pruning of the given tree. The algorithm proceeds to prune the nodes of a branch as long as both subtrees of an internal node are pruned and stops immediately if even one subtree is kept. Moreover, it prunes an interior node only if all its descendants at level $d$ have been pruned. Furthermore, REP either halts before the safe nodes are reached or prunes the whole tree only in case all safe nodes have the same majority class.

In the second setting the tree under consideration was assumed to fit noise; i.e., it was assumed that the class label of the pruning examples is independent of their attribute values. In this setting the pruning probability of the tree could be bound by an equation that depends exponentially on the size of the tree and linearly on the number and class distribution of the pruning examples. Thus, our analysis corroborates the main finding of Oates and Jensen (1999) that REP fails to control the growth of a decision tree in the extreme case that the tree fits pure noise. Moreover, our analysis opened a possibility to initially explain why the learned decision tree grows linearly with an increasing data set. Our bound on the pruning probability of a tree is based on bounding the probability that all safe nodes have the same majority class. Surprisingly, essentially the same property, whose probability we try to bound close to 0, is assumed to hold with probability 1 in the analysis of Oates and Jensen (1999).

In REP it may happen that no pruning examples are directed to a given subtree. Such subtrees have not been taken into account in earlier analyses. In our final analysis we included empty subtrees in the equation for a tree's pruning probability. Taking empty subtrees into account gives a more realistic bound for the pruning probability of a tree.

Unfortunately, one cannot draw very definite general conclusions on the two-phased top-down induction of decision trees on the basis of analyses on the REP algorithm, because its bias is quite unique among pruning algorithms. The fact that REP does not penalize the size of a tree, but only rests on the classification error on the pruning examples makes the method sensitive to small changes in the class distribution of the pruning set. Other decision tree pruning algorithms also have their individual characteristics. Therefore, unified analysis of decision tree pruning may be impossible.

The version of REP, in which one is allowed to relabel original leaves, as well, is used as the performance objective in Kearns and Mansour's (1998) pruning algorithm. Thus, the performance of pruning algorithms that use both error and size penalty is related to those that use only error estimation. In the version of REP used by Kearns and Mansour our analysis based on safe nodes applies with leaves in place of safe nodes. Hence for this algorithm the derived bounds are stricter.

We leave the detailed analysis of other important pruning algorithms as future work. Only through such investigation is it possible to disclose the differences and similarities of pruning algorithms. Empirical examination has not managed to reveal clear performance differences between the methods. Also, the relationship of the number of safe nodes and leaves of a tree ought to be examined analytically and empirically. In particular, one should study whether the number of safe nodes does increase linearly with a growing training set, as conjectured in this paper. Deeper understanding of existing pruning algorithms may help to overcome the problems associated with the pruning phase of decision tree learning.